\documentclass[runningheads]{llncs}
\usepackage{graphicx}
\usepackage{amsmath}
\usepackage{amssymb}
\usepackage{amsmath}
\usepackage{algorithm}
\usepackage{algpseudocode}

\DeclareMathOperator*{\argmin}{arg\,min}
\begin{document}
\title{Unsupervised 4D Cardiac Motion Tracking with Spatiotemporal Optical Flow Networks}
\author{Long Teng\inst{1,2,3}\and
wei.feng\inst{1,2,4} \and
menglong.zhu\inst{3} \and
xinchao.li\inst{3} 
}

\institute{Shenzhen Institute of Advanced Technology, Chinese Academy of Sciences, Shenzhen 518055, China\\ \and
University of Chinese Academy of Sciences, Beijing 100049, China\\ \and
DJI Technology Co., Ltd., DJI Sky City, No. 55 Xianyuan Road, Nanshan District, Shenzhen, China\\ \and
Shenzhen University of Advanced Technology\\ 
}
\maketitle              % typeset the header of the contribution
\begin{abstract}
Cardiac motion tracking from echocardiography can be used to estimate and quantify myocardial motion within a cardiac cycle. It is a cost-efficient and effective approach for assessing myocardial function. However, ultrasound imaging has the inherent characteristics of spatially low resolution and  temporally random noise, which leads to difficulties in obtaining reliable annotation. Thus it is difficult to perform supervised learning for motion tracking. In addition, there is no end-to-end unsupervised method currently in the literature. This paper presents a motion tracking method where unsupervised optical flow networks are designed with spatial reconstruction loss and  temporal-consistency loss. Our proposed loss functions make use of the pair-wise and temporal correlation to estimate cardiac motion from noisy background. Experiments using a synthetic 4D echocardiography dataset has shown the effectiveness of our approach, and its superiority over existing methods on both accuracy and running speed. To the best of our knowledge, this is the first work performed that uses unsupervised end-to-end deep learning optical flow network for 4D cardiac motion tracking.

\end{abstract}
\section{Introduction}
4D motion tracking is used to describe and quantify how the myocardial wall moves in 3D sequence data. It is a fundamental analysis of 3D image sequences, which also helps other tasks like segmentation, classification, and abnormal detection. Other than pair-wise 3D image registration or optical flow estimation, 4D motion tracking depends on both pair-wise constraint and sequence correlation to estimate the possible motion fields.

Several approaches have been proposed to approximate motion tracking\cite{de2012temporal,parajuli2016integrated,parajuli2019flow}. De Craene et al. proposed a 3D+t diffeomorphic based registration to estimate the motion fields \cite{de2012temporal}. Their registration uses a b-spline parameterization over the velocity field, but its Lagrangian displacements are prone to accumulate error by integrating the velocities. Nripesh et al. proposed a dynamic programming method to accomplish patch matching. However, they did not make use of the spatial constraints \cite{parajuli2016integrated}. A recent improvement from Nripesh et al. uses the Siamese similarity as the weights of dynamic programming to construct spatiotemporal constraints \cite{parajuli2019flow}. Although this work makes uses of spatial constraints, their dynamic programming still relies on the segmentation result. To summarize, the key limitations of existing methods for 4D cardiac motion tracking are: 1) considering only spatial or temporal correlation  \cite{parajuli2016integrated}, 2)extra errors from each  of multi-stage method \cite{parajuli2019flow}, and 3) error accumulation when integrate the velocities over time \cite{de2012temporal}. 

Recent research efforts have shown the possibility of using a Convolutional Neural Network (CNN) based optical flow estimation based on image processing \cite{dosovitskiy2015flownet,fan2018end,zach2007duality} and can potentially overcome the limitations of the existing methods for 4D cardiac motion tracking as we state above. Dosovitskiy et al. proposed a Flow-Net which stacks two images as input and output the flow field in cascade multi-scale \cite{dosovitskiy2015flownet}.  Later, Fan et al. considered temporal correlation and proposed a TVNet \cite{fan2018end}, which is based on TVL1 \cite{zach2007duality} to classify videos using optical flow estimation. However, Flow-Net \cite{dosovitskiy2015flownet} was trained on large scale synthetic images with ground truth labels, which is infeasible for medical data analysis, where we usually do not have ground truth annotation for tracking. TVNet \cite{fan2018end} is promising, but it was designed for 2D image sequence and directly extend the 3D version on 4D cardiac motion tracking problem would be problematic on computational memory, hence hurt model performance .

To overcome those limitations, we make use of 4D spatiotemporal information in designing the CNN model and loss function (shown in figure \ref{fig:1}). From the model aspect, a trainable optical flow is used as a guideline to estimate motion fields. From the loss function aspect, temporal-consistency is used for leveraging the temporal correlation, and reconstruction loss is used for leveraging the spatial correlation. A reference frame is used to reduce the accumulation error. The contributions of this paper, therefore, can be summarized in two ways:
\begin{itemize}
\setlength{\itemsep}{0pt}
\setlength{\parsep}{0pt}
\setlength{\parskip}{0pt}
\item A novel, 3D unsupervised optical flow network to estimate motion fields was developed.

\item Different than existing methods, we simultaneously invoke constraints on temporal-consistency and spatial reconstruction by using novel losses to leverage the spatiotemporal correlation. 
\end{itemize}

%-------------------------------------------------------------------------

\section{\label{section3}Model Architecture}
\begin{figure*}
    \setlength{\abovecaptionskip}{0.cm}
	\setlength{\belowcaptionskip}{-0.cm}
    \begin{center}
        \includegraphics[scale=0.35]{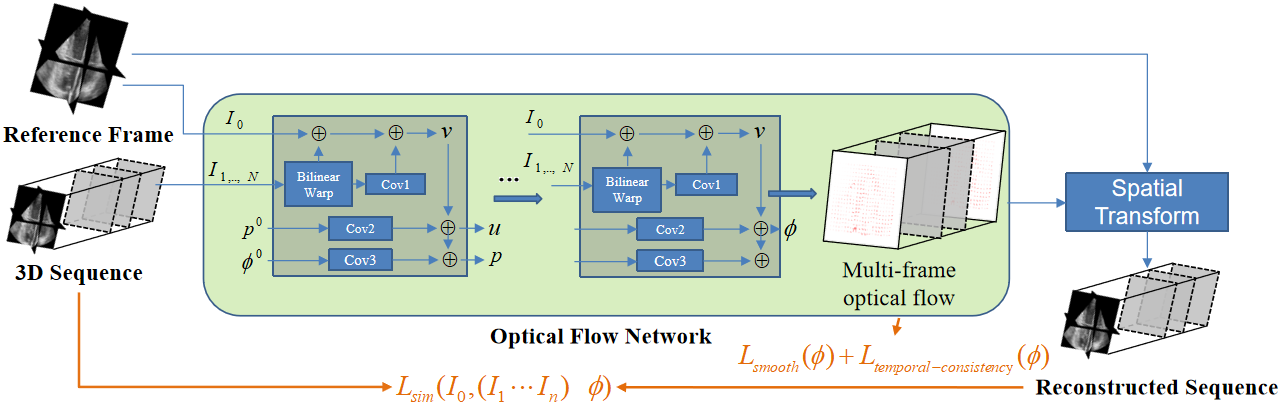}
        \caption{\label{fig:1} Overview of our Cardiac Motion Tracking with Spatialtemporal Networks framework: The optical flow network takes the reference frame, and an image sequence as input and outputs reconstructed sequence. Three loss functions are illustrated with the orange arrows and equations. Notice that, the 3D sequence contains both forward time sequence and reverse time sequence will be discussed in section \ref{section4}.}
    \end{center}
\end{figure*}

Among previous optical flow network architectures, we use the TV-L1 based networks and extend it into 3D \cite{guo2010action}. The benefit of this network is the minimal requirement of the training dataset. To derive the formulation of 3D networks, we begin from the original TV-L1 with L1 norm Horn and Schunck's equation\cite{horn1981determining}.
\begin{equation}
    \begin{aligned}
    \label{eq:3}
    \hat{\phi}_i&=\argmin_{\phi_i}\mathcal{L}(I_0,I_i,\phi_i)  
    =\argmin_{\phi_i}\lambda\mathcal{L}_{sim}(I_0,I_i,\phi_i)  +  \mathcal{L}_{smooth}(\phi_i) \\
    &=\argmin_{\phi_i} \!\lambda\! \!\int_{\Omega}\!|\rho(\phi_i)|d\Omega\!+\! \!\int_{\Omega}\!|\nabla \phi_i^x| \!+\! |\nabla \phi_i^y|\!+\!|\nabla \phi_i^z|d\Omega\\
    \end{aligned}
\end{equation}
where $I_0$ is the reference frame, $I_i$ is the $i$th frame in the sequence with length $n$ and $i \in \{1,\dots, n\}$. $\phi_i = [\phi_i^x,\phi_i^y,\phi_i^z]^T$ represents the motion component toward three coordinate directions. $\rho(\phi_i)=I_0-m(I_i,\phi_i)$ is the similarity between inference frame and a specific frame in the sequence. With the help of fixed-point iteration, Eq. \eqref{eq:3} can be solved by iteration of Eq. \eqref{eq:8} and Eq. \eqref{eq:9} \cite{fan2018end,perez2013tv}. First, when auxiliary variable $v$ is fixed, we update motion field $\phi_i$ and the vector field $p$ that is initialized to minimize $\phi_i$ with
% \begin{equation}
% \label{eq:8}
% \left\{\begin{array}{ll}
%       \phi = v + \theta \cdot divp &\\
%       p = \frac{p+\tau/\theta\cdot\nabla\phi}{1+\tau/\theta\cdot|\nabla\phi|}&
% \end{array}\right.
% \end{equation}

\begin{equation}
\label{eq:8}
\phi_i = v + \theta \cdot divp; \quad
p = \frac{p+\tau/\theta\cdot\nabla\phi_i}{1+\tau/\theta\cdot|\nabla\phi_i|},
\end{equation}
where $\tau$ is the time-step, $div$ is the operation of divergence. $\theta$ is the weight in regularization term. Then $\phi_i$ is fixed, we update $v$ by the optimal solution:
\begin{equation}
\label{eq:9}
v = \left
\{\begin{array}{ll}
       \lambda\theta\nabla I_i, \: \: \: &if \: \rho(\phi_i)<-\lambda\theta|\nabla I_i|^2 \\
       -\lambda\theta\nabla I_i, \: \: \: &if \: \rho(\phi_i)>\lambda\theta|\nabla I_i|^2 \\
       -\frac{\rho(\phi_i)\nabla I_i}{|\nabla I_i|^2}, \: \: \: &if \: |\rho(\phi_i)|\leq\lambda\theta|\nabla I_i|^2
\end{array}\right.
\end{equation}
where $\nabla I_i=\nabla I_i(x+\phi_i)$ for short.

As it is shown in Fig. \ref{fig:1}, the base module of optical flow networks contains only 9 convolution layers and a 3D warp layer. The convolution blocks in figure \ref{fig:1} are marked as Conv1, Conv2 and Conv3. Each block contains three convolution layers in three directions. They are trainable and initialized by,
\begin{equation}
\begin{aligned}
\label{eq:10}
\nabla I_i &= \frac{\partial I_i(x,y,z)}{\partial x} + \frac{\partial I_i(x,y,z)}{\partial y} + \frac{\partial I_i(x,y,z)}{\partial z} \\
\end{aligned}
\end{equation}
The optical flow estimation process are described in Algorithm 1. 
\begin{table}
\begin{center}
\begin{tabular}{l}
\hline
\textbf{Algorithm 1}: Optical flow estimation process\\
\hline
{\textbf{Pre-defined Hyper-parameters}: $\lambda$, $\theta$, $N_{iters}$ } \\
%\textbf{Initial Hyper-parameters}: $\lambda$, $\theta$, $N_{iters}$  \\
\textbf{Initial Convolution Weights}:$I_0$, $I_1$, $\phi= 0$    \\
\:\:\:\:\:$w_{Conv1_j} \gets [[-0.5, 0, 0.5]]$  \:\:\:\:\:\:$w_{Conv2_j} \gets [[[-1, 1]]]$     \:\:\:\:\:\:\:$w_{Conv3_j} \gets [[[-1, 1]]], \forall  j \in \{x,y,z\}$ \\
\:\:\:\:\:$p_{x} = [p_{xx}, p_{xy}, p_{xz}] \gets [0,0,0]$ \:\:\:\:\:$p_{y} = [p_{yz}, p_{yy}, p_{yz}] \gets [0,0,0]$\\ \:\:\:\:\:$p_{z} = [p_{zx}, p_{zy}, p_{zz}] \gets [0,0,0]$
\:\:\:\:\:$\rho(\phi_1)\gets I_1(x+\phi_0) + (\phi_1-\phi_0)\nabla I_1(x+\phi_0) - I_0(x)$ \\
\textbf{Input}: $I_0$, $I_i(i \in \{1,...,n\})$ \\
\textbf{for} $N_{iters}$ interations \textbf{do} \\
 \:\:\:\:\:
$v \gets \left
\{\begin{array}{ll}
      \lambda\theta\nabla I_i, \: \: \: &if \: \rho(\phi_i)<-\lambda\theta|\nabla I_i|^2 \\
      -\lambda\theta\nabla I_i, \: \: \: &if \: \rho(\phi_i)>\lambda\theta|\nabla I_i|^2 \\
      -\frac{\rho(\phi_i)\nabla I_i}{|\nabla I_i|^2}, \: \: \: &if \: |\rho(\phi)|\leq\lambda\theta|\nabla I_i|^2
\end{array}\right.$\\
\:\:\:\:\:where $\nabla I_i=\nabla I_i(x+\phi_i)$ for short; \\
\:\:\:\:\:  $\phi_{i\_j} \gets v + \theta \cdot divp_j$ \:\:\:\:\:  $p_j \gets \frac{p+\tau/\theta\cdot\nabla\phi_i}{1+\tau/\theta\cdot|\nabla\phi_i|},\forall  j \in \{x,y,z\}$\\
\textbf{end} \\
\textbf{Output: $I_i(x+\hat{\phi_i})$, $\hat{\phi_i}$}\\
\hline
\end{tabular}
\end{center}
\end{table}

\section{\label{section4}Loss Function}
Loss function is important to the deep learning based methods. In this paper, we propose to use both spatial constraint and temporal constraints to estimate the desired motion tracking from optical flow. As is shown in figure \ref{fig:1}, the orange lines and equations represent the loss functions. The final loss function contains three loss functions. Those functions are divided into two categories, temporal-consistency loss, and spatial-reconstruction loss.

\subsection{Temporal-Consistency Loss\label{sec:4.1}}
The myocardium moves periodically in echocardiography while the noise randomly appears. This physical meaning indicates that the desired motion fields meet the temporal consistency.
\begin{figure*}[b]
    \setlength{\abovecaptionskip}{0.cm}
	\setlength{\belowcaptionskip}{-0.cm}
    \begin{center}
        \includegraphics[scale=0.38]{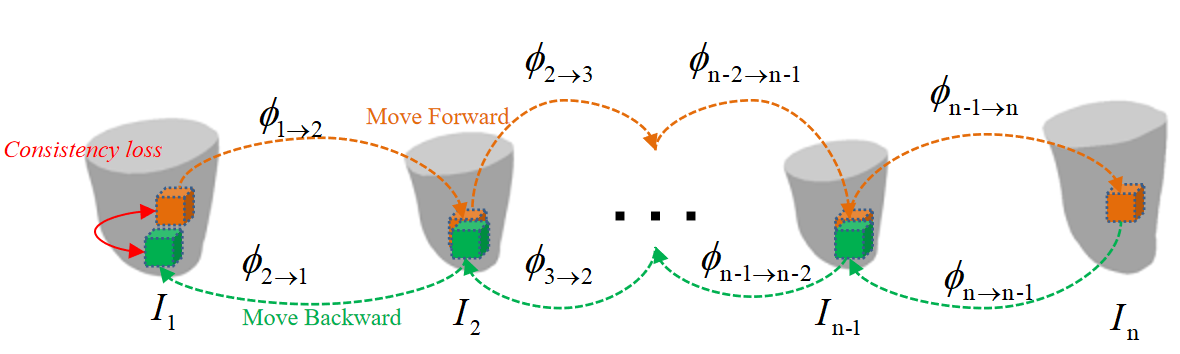}
        \caption{\label{fig:3} The Temporal-Consistency loss for motion estimation.}
    \end{center}
\end{figure*}

As it is shown in figure \ref{fig:3}, $\{I_1 \cdots I_n\}$ represent the 3D echocardiography sequence. The optical flow field $\phi_{i\xrightarrow{} i+1}$ means the optical flow from $I_i$ to $I_{i+1}$. Correspondingly, $\phi_{i\xrightarrow{} i-1}$ is the optical flow from $I_i$ to $I_{i-1}$. Every point in $I_1$ changes its position from frame to frame by interpolation with the optical flow of $\phi$. The forward interpolation loop is defined as follows,
\begin{equation}
\begin{aligned}
\label{eq:11}
f_{wd}(\phi) = \textsl{h}(\textsl{h}(\phi_{1\xrightarrow{}2}, \phi_{1\xrightarrow{}2}),\cdots,\phi_{n-1\xrightarrow{}n})
\end{aligned}
\end{equation}
where function $\textsl{h}$ represents interpolation operation that maps images by the given optical flow. And the backward interpolation loop is,
\begin{equation}
\begin{aligned}
\label{eq:12}
b_{wd}(\phi) = \textsl{h}(\textsl{h}(\phi_{n\xrightarrow{}n-1}, \phi_{n\xrightarrow{}n-1}),\cdots,\phi_{2\xrightarrow{}1})
\end{aligned}
\end{equation}
As it is illustrated in figure \ref{fig:3}, a selected point in $I_1$ has meshed in orange color. Its position changes over time by the interpolation with $\phi$. When it moves forward to $I_n$ and then moves back to $I_1$, it may not move back to its original position. There are two reasons for this error. Firstly, errors exist in the pairwise optical flow estimate. These errors accumulate throughout time and lead to temporal inconsistency. Secondly, the inherent low resolution and random noise of echocardiography also lead to this temporal inconsistency.
To minimize the temporal inconsistency, Temporal-Consistency loss is introduced as follows,
\begin{equation}
\begin{aligned}
\label{eq:13}
L_{Cycle}(\phi) = |\phi_{1\xrightarrow{}2} - b_{wd}(f_{wd}(\phi))|
\end{aligned}
\end{equation}
The final Temporal-Consistency loss is the combination of whole cycle loss and single cycle loss,
\begin{equation}
\begin{aligned}
\label{eq:15}
L_{Temporal\_Consistency} = L_{Sig}(\phi) + \omega L_{Cycle}(\phi)
\end{aligned}
\end{equation}
where $L_{Sig}(\phi) = \frac{1}{n}\sum_{i=1}^{n}|\phi_{i\xrightarrow{}i+1} + m(\phi_{i+1\rightarrow{} i}, \phi_{i+1\xrightarrow{}i})|$ is the single cycle loss.

\subsection{Spatial Reconstruction Loss}
The spatial reconstruction loss contains two parts. First part is the similarity between original $I_0$ and the warped $I_1$ with respect to $\phi$. Second part is the smoothness function of optical flow $\phi$. The loss function is similar to Eq. \eqref{eq:3}. The overall learning objective sums combines temporal loss and spatial loss,
\begin{equation}
    \begin{aligned}
    \label{eq:17}
    L_{ST} = \gamma L_{Temporal\_Consistency} + \beta L_{rec}
    \end{aligned}
\end{equation}
Where the spatial loss function is $L_{rec} = \lambda\! \!\int_{\Omega}\!|\rho(\phi)|d\Omega\!+\! \!\int_{\Omega}\!|\nabla \phi_x| \!+\! |\nabla \phi_y|\!+\!|\nabla \phi_z|d\Omega$. 

%-------------------------------------------------------------------------
\section{Experiment\label{section5}}

\subsection{Dataset}
The data we used to conduct the experiment was from an open-access dataset, 3D Strain Assessment in Ultrasound(STRAUS) \cite{alessandrini2015pipeline}. It contains 8 groups of B-mode simulated voxel data of ischemic sequences with ischemia in the distal  (LADDIST), proximal left anterior descending artery (LADPROX), left bundle branch block (LBBB and LBBBSMALL), left circumflex artery (LCX),  right circumflex artery (RCA), synchronous (SYNC) and a normal group (NORMAL). Each group has 34 to 41 frames of 3D images. And there are 286 frames in total.

\subsection{Implementation Details}
All experiments were accomplished with GTX1080 GPU. The original image had $224\times176\times208$ voxels of size $0.7\times0.9\times0.6 mm^3$.  We performed random shifting, flipping, and cropping in three coordinate directions for the data augmentation. The input 3D images were resized to $64\times64\times64$ with bilinear interpolation. To compare with ground truth, the outputs were resized back to the original size. The scaling is a trade-off between accuracy and GPU memory limits.

The iteration number of optical flow network $N_{iters}$ is 40.  The temporal sequence length is 4. The parameters in Eq. \eqref{eq:17}, $\gamma$, $\omega$, $\beta$, $\lambda$ are 1.0, 1.0, 0.5 and 0.15 representatively. The illustrated resulting frame is randomly selected throughout the cardiac cycle.

\subsection{Comparison with ground truth\label{chap:4.3}}

\begin{minipage}[t]{\textwidth}
 \begin{minipage}[t]{0.45\textwidth}
    \centering
    \begin{tabular}{ccc} 
	\hline
    Data& MSE(mm) & EPE\\
    \hline
    LADPROX& $0.38\pm 0.31$ & $0.78\pm 0.74$\\
    LBBB& $0.38\pm 0.40$ & $0.88\pm 0.73$\\
    LADDIST& $0.45\pm 0.43$ & $0.95\pm 0.79$\\
    \hline
	\end{tabular}
	\makeatletter\def\@captype{table}\makeatother\caption{\label{tab:1}Results of MSE and EPE on three testing dataset}
  \end{minipage}
  \begin{minipage}[t]{0.45\textwidth}
    \centering
    \begin{tabular}{cc}        
    \hline
    Methods& MSE(mm)\\
    \hline
    BSR& $1.66\pm 1.05$\\
    DST& $1.23\pm 0.89$\\
    GRPM& $1.21\pm 1.13$\\
    FNT& $1.10\pm 0.74$\\
    \textbf{Ours}(No temporal loss)& $\textbf{0.42}\pm \textbf{0.52}$\\
    \textbf{Ours}(With temporal loss)& $\textbf{0.41}\pm \textbf{0.47}$\\
    \hline
	  \end{tabular}
	  \makeatletter\def\@captype{table}\makeatother\caption{\label{tab:2}MSE for related motion tracking methods and our result.}
   \end{minipage}
\end{minipage}

\begin{figure*}[t]
    \setlength{\abovecaptionskip}{0.cm}
	\setlength{\belowcaptionskip}{-0.cm}
    \begin{center}
        \includegraphics[scale=0.45]{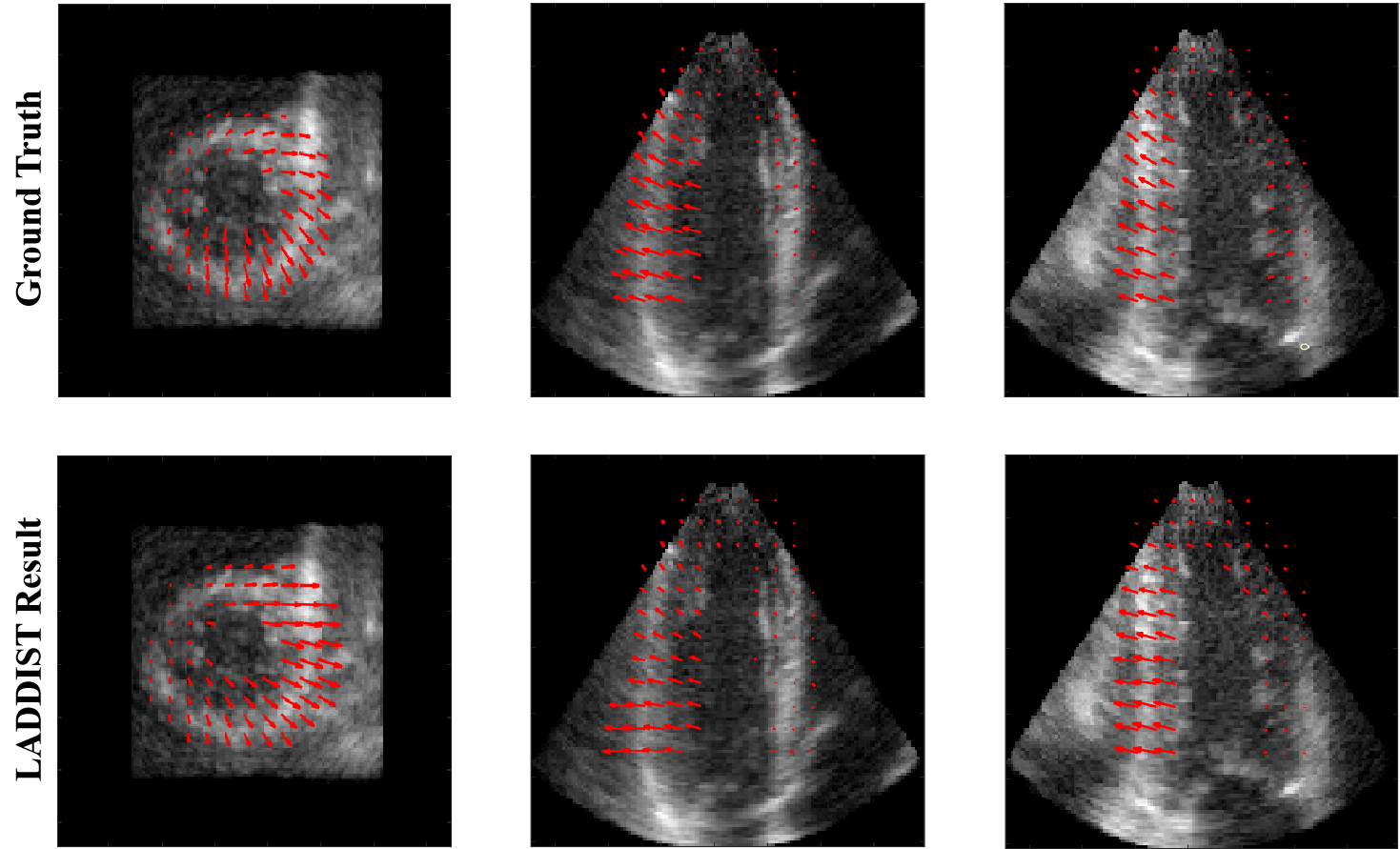}
        \caption{\label{fig:4} \textbf{LADDIST}: Illustrated frame is the 30th of 34 frames. Heterogeneous motion caused by ischemia is captured. }
    \end{center}
\end{figure*}
Three groups of data (LADDIST,  LBBB,  LADDIST) with 106 frames of 3D images were used for testing. The rest 180 frames were used for training.  We applied the Average Mean Square Error(MSE) and Average End-Point Error(EPE) as the criterions for comparison. Results are listed in tabel \ref{tab:1},

Figure \ref{fig:4} is the result slice of LADDIST in $x$, $y$, $z$ directions.  The first row is the ground truth, the second row is our result. The motion field are sampled into a grid at intervals of $7$. Only the myocardium data is used to estimate the cardiac motion. The ground truth of the motion field and our result both show the same heterogeneous motion patterns. Normal heart contract towards the heart center, while ischemia in the distal lead to abnormal diastole. The estimated motion in Figure \ref{fig:4} captures the abnormal motions. This could help to the diagnosis. 

\begin{figure*}[t]
    \setlength{\abovecaptionskip}{0.cm}
	\setlength{\belowcaptionskip}{-0.cm}
    \begin{center}
        \includegraphics[scale=0.45]{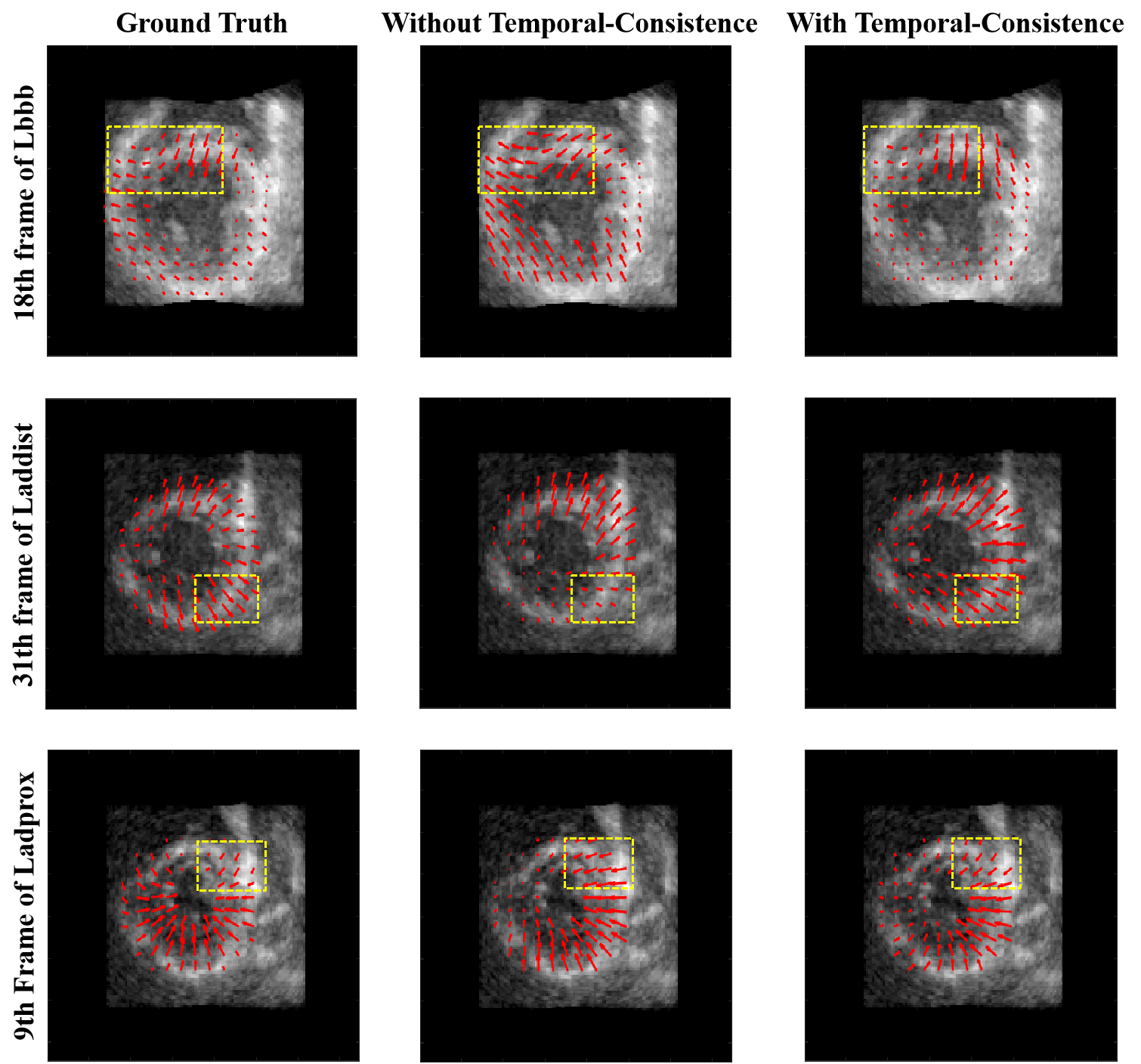}
        \caption{\label{fig:7} \textbf{LBBB, LADDIST, LADPROX}: Randomly selected results from three testing datasets. The comparison of the motion field with and without training with the Temporal-Consistency loss function.}
    \end{center}
\end{figure*}
\subsection{Ablation study of Temporal-Consistency}
Reconstruction loss is the basic loss for regulation, we only perform ablation study on temporal loss. The MSE index with Temporal-Consistency loss improves $0.01\pm 0.05$ in the three dataset. The estimation error is greatest around the 10th frame and smallest in the first and last frames. Although the MSE index improvement is not obvious, it leads to visually more accurate details.

Figure \ref{fig:7} gives randomly selected frames from three datasets. The three dataset contains different types of heart disease. After training with Temporal-Consistency loss, the motion field captures more accurate heterogenours motion to the ground truth. As we can see in figure \ref{fig:7}, although the results with Temporal-Consistence are similar to the result without Temporal-Consistence (consistent with limited MSE improvement shown in Table \ref{tab:2}), it provides more accurate details. Due to the limitation of GPU memory, we only use 4 frames in Temproal-Consistency. We noticed that the descrenpcies in the middel frames were larger (shown in supplymentary). As we can imagining, with more temporal length as input for training, the improvement of using temporal consistency loss will be enlarged.

\subsection{Comparison with other methods}
We compared our result with the recently published results of 4 methods \cite{parajuli2017flow}, B-spline based free
form deformation registration (BSR) \cite{B-spline}, the dynamic shape tracking (DST) \cite{parajuli2016integrated}, a simplified version of the generalized robust point matching algorithm (GRPM) \cite{lin2004generalized} and flow net motion tracking (FNT) \cite{parajuli2017flow}. The results were listed in table \ref{tab:2}.

Our results significantly outperformed other methods and was almost half of their errors. Beside, our average running speed is $0.3s$ per frame on GTX1080 GPU which was promising for clinical application.

\section{Conclusion and future work\label{section6}}
In this paper, we proposed a novel end-to-end unsupervised optical flow network for 4D cardiac motion tracking. Our model leverages the spatio-temporal correlation by using the reconstruction loss and temporal-consistency loss. We validate our method quantitatively and qualitatively on three synthetic 4D echocardiography datasets. The presented networks outperformed the alternative methods in terms of accuracy and running speed (0.3s per frame). Most importantly, our method does not require ground truth for training, and thus overcomes the annotation limitation. 
 To the best of our knowledge, this is the first effort that uses unsupervised end-to-end deep learning on 4D cardiac motion tracking. Our method shows the potential of applying to real data. However, real data usually only has limited sparse key points as the ground truth for motion tracking\cite{tobon2013benchmarking} and lacks fair evaluation metrics for motion tracking results. Our future work is to adapt the proposed algorithm on real data and propose suitable evaluation metrics. 
%\bibitem{ref_lncs0}
%Author, F.: Article title. Journal \textbf{2}(5), 99--110 (2016)

%Author, F., Author, S.: Title of a proceedings paper. In: Editor,
%F., Editor, S. (eds.) CONFERENCE 2016, LNCS, vol. 9999, pp. 1--13.
%Springer, Heidelberg (2016). \doi{10.10007/1234567890}

%Author, F., Author, S., Author, T.: Book title. 2nd edn. Publisher,

%Author, A.-B.: Contribution title. In: 9th International Proceedings
%on Proceedings, pp. 1--2. Publisher, Location (2010)

%LNCS Homepage, \url{http://www.springer.com/lncs}. Last accessed 4
%\end{thebibliography}
\newpage
{\small
\bibliographystyle{unsrt}
\bibliography{egbib}

\begin{thebibliography}{10}

\bibitem{de2012temporal}
Mathieu De~Craene, Gemma Piella, Oscar Camara, Nicolas Duchateau, Etelvino Silva, Adelina Doltra, Jan D¡¯hooge, Josep Brugada, Marta Sitges, and Alejandro~F Frangi.
\newblock Temporal diffeomorphic free-form deformation: Application to motion and strain estimation from 3d echocardiography.
\newblock {\em Medical image analysis}, 16(2):427--450, 2012.

\bibitem{parajuli2016integrated}
Nripesh Parajuli, Allen Lu, John~C Stendahl, Maria Zontak, Nabil Boutagy, Melissa Eberle, Imran Alkhalil, Matthew O¡¯Donnell, Albert~J Sinusas, and James~S Duncan.
\newblock Integrated dynamic shape tracking and rf speckle tracking for cardiac motion analysis.
\newblock In {\em International Conference on Medical Image Computing and Computer-Assisted Intervention}, pages 431--438. Springer, 2016.

\bibitem{parajuli2019flow}
Nripesh Parajuli, Allen Lu, Kevinminh Ta, John Stendahl, Nabil Boutagy, Imran Alkhalil, Melissa Eberle, Geng-Shi Jeng, Maria Zontak, Matthew O¡¯Donnell, et~al.
\newblock Flow network tracking for spatiotemporal and periodic point matching: Applied to cardiac motion analysis.
\newblock {\em Medical image analysis}, 55:116--135, 2019.

\bibitem{dosovitskiy2015flownet}
Alexey Dosovitskiy, Philipp Fischer, Eddy Ilg, Philip Hausser, Caner Hazirbas, Vladimir Golkov, Patrick Van Der~Smagt, Daniel Cremers, and Thomas Brox.
\newblock Flownet: Learning optical flow with convolutional networks.
\newblock In {\em Proceedings of the IEEE international conference on computer vision}, pages 2758--2766, 2015.

\bibitem{fan2018end}
Lijie Fan, Wenbing Huang, Chuang Gan, Stefano Ermon, Boqing Gong, and Junzhou Huang.
\newblock End-to-end learning of motion representation for video understanding.
\newblock In {\em Proceedings of the IEEE Conference on Computer Vision and Pattern Recognition}, pages 6016--6025, 2018.

\bibitem{zach2007duality}
Christopher Zach, Thomas Pock, and Horst Bischof.
\newblock A duality based approach for realtime tv-l 1 optical flow.
\newblock In {\em Joint pattern recognition symposium}, pages 214--223. Springer, 2007.

\bibitem{guo2010action}
Kai Guo, Prakash Ishwar, and Janusz Konrad.
\newblock Action recognition using sparse representation on covariance manifolds of optical flow.
\newblock In {\em 2010 7th IEEE international conference on advanced video and signal based surveillance}, pages 188--195. IEEE, 2010.

\bibitem{horn1981determining}
Berthold~KP Horn and Brian~G Schunck.
\newblock Determining optical flow.
\newblock {\em Artificial intelligence}, 17(1-3):185--203, 1981.

\bibitem{perez2013tv}
Javier~S{\'a}nchez P{\'e}rez, Enric Meinhardt-Llopis, and Gabriele Facciolo.
\newblock Tv-l1 optical flow estimation.
\newblock {\em Image Processing On Line}, 2013:137--150, 2013.

\bibitem{alessandrini2015pipeline}
Martino Alessandrini, Mathieu De~Craene, Olivier Bernard, Sophie Giffard-Roisin, Pascal Allain, Irina Waechter-Stehle, J{\"u}rgen Weese, Eric Saloux, Herv{\'e} Delingette, Maxime Sermesant, et~al.
\newblock A pipeline for the generation of realistic 3d synthetic echocardiographic sequences: Methodology and open-access database.
\newblock {\em IEEE transactions on medical imaging}, 34(7):1436--1451, 2015.

\bibitem{parajuli2017flow}
Nripesh Parajuli, Allen Lu, John~C Stendahl, Maria Zontak, Nabil Boutagy, Imran Alkhalil, Melissa Eberle, Ben~A Lin, Matthew O¡¯Donnell, Albert~J Sinusas, et~al.
\newblock Flow network based cardiac motion tracking leveraging learned feature matching.
\newblock In {\em International Conference on Medical Image Computing and Computer-Assisted Intervention}, pages 279--286. Springer, 2017.

\bibitem{B-spline}
Dirk-Jan Kroon.
\newblock B-spline grid, image and point based registration.
\newblock 2008.

\bibitem{lin2004generalized}
Ning Lin and James~S Duncan.
\newblock Generalized robust point matching using an extended free-form deformation model: Application to cardiac images.
\newblock In {\em 2004 2nd IEEE International Symposium on Biomedical Imaging: Nano to Macro (IEEE Cat No. 04EX821)}, pages 320--323. IEEE, 2004.

\bibitem{tobon2013benchmarking}
Catalina Tobon-Gomez, Mathieu De~Craene, Kristin Mcleod, Lennart Tautz, Wenzhe Shi, Anja Hennemuth, Adityo Prakosa, Hengui Wang, Gerry Carr-White, Stam Kapetanakis, et~al.
\newblock Benchmarking framework for myocardial tracking and deformation algorithms: An open access database.
\newblock {\em Medical image analysis}, 17(6):632--648, 2013.

\end{thebibliography}
}

\end{document}